\documentclass{article}

\usepackage[final,nonatbib]{neurips_2021}

\usepackage[utf8]{inputenc} 
\usepackage[T1]{fontenc}    
\usepackage{hyperref}       
\usepackage{url}            
\usepackage{booktabs}       
\usepackage{amsfonts}       
\usepackage{nicefrac}       
\usepackage{microtype}      
\usepackage{xcolor}         

\usepackage{times}
\usepackage{latexsym}
\usepackage{booktabs}
\usepackage{caption}
\usepackage{graphicx}
\usepackage{amsmath}
\usepackage{amsfonts}
\usepackage{siunitx}
\usepackage{subcaption}
\usepackage{amssymb}
\usepackage{tabularx}
\usepackage{multirow}
\usepackage{colortbl}
\definecolor{Gray}{gray}{0.9}
\definecolor{Gray2}{gray}{0.8}

\title{Efficient Variational Graph Autoencoders for Unsupervised Cross-domain Prerequisite Chains}

\author{%
  Irene Li, Vanessa Yan, Dragomir Radev \\
  Department of Computer Science\\
  Yale University\\
  \texttt{\{irene.li,vanessa.yan,dragomir.radev\}@yale.edu}
}

\begin{document}

\maketitle

\begin{abstract}
  
Prerequisite chain learning helps people acquire new knowledge efficiently. While people may quickly determine learning paths over concepts in a domain, finding such paths in other domains can be challenging. We introduce Domain-Adversarial Variational Graph Autoencoders (DAVGAE) to solve this cross-domain prerequisite chain learning task efficiently. Our novel model consists of a variational graph autoencoder (VGAE) and a domain discriminator. The VGAE is trained to predict concept relations through link prediction, while the domain discriminator takes both source and target domain data as input and is trained to predict target domain labels. Most importantly, this method only needs simple homogeneous graphs as input, unlike the current state-of-the-art model which requires the construction of heterogeneous graphs. We evaluate our model on the LectureBankCD dataset, and results show that our model outperforms recent graph-based benchmarks while using only 1/10 of graph scale and 1/3 of computation time.
\end{abstract}

\section{Introduction}
A prerequisite is defined as a concept which must be learned prior to another concept. Knowing prerequisite relationships between concepts helps determine learning paths for students who wish to acquire new knowledge. \cite{alsaad2018mining,li2019whatsi, yu2020mooccube}. Most existing work on prerequisite chain learning is limited to a single domain. \cite{yang2015concept,pan-etal-2017-prerequisite,liang2018investigating}. More recently, \cite{li2021unsupervised} introduced the task of cross-domain prerequisite chain learning.
This task is useful for development of educational resources, intelligent search engine rankings, and other services for people who may have an excellent background in one domain, such as natural language processing (NLP), and wish to learn concepts in a new domain, such as Bioinformatics (BIO).
These two domains share common fundamental concepts: machine learning basics, time-series data, statistics, etc. Given the prerequisite chains in a source domain, it is possible to transfer the knowledge to learn prerequisite chains in a target domain.  
\cite{li2021unsupervised} proposed the CD-VGAE (cross-domain variational graph autoencoder) model to apply domain transfer and infer target concept relations.

However, CD-VGAE was trained on a complex graph that contains many resource nodes and concept nodes from both the source and target domains, making it limited in scalability. A known challenge of applying graph neural networks in practice is the difficulty of scaling these models to large graphs. \cite{bojchevski2020scaling, frasca2020SIGN}. We seek to develop a model that can be trained on a much smaller graph than the CD-VGAE model, so that the model can be more practical in real-world applications. Specifically, our model is trained on graphs with concept nodes only.

Adversarial methods \cite{Goodfellow2014generative} have been applied to NLP tasks that involve multilingual or multi-domain scenarios \cite{chen2018adversarial,chen2018multinomial,li2021detecting}. Such methods typically introduce a domain loss to a neural network in order to learn domain-invariant features for unsupervised domain adaptation. However, there has been limited research in training adversarial networks on graphs. The only existing work is the adversarially regularized variational graph autoencoder (ARVGA) model \cite{pan2018adversarially} , which learns robust graph embeddings by reconstructing graph structure. We introduce a variant adversarial framework to solve cross-domain prerequisite chain learning.

Our contributions are two-fold. First, we propose domain-adversarial variational graph autoencoders (DAVGAE) to perform unsupervised cross-domain prerequisite chain learning. Second, we offer two ways to construct the concept graph: cross-domain and single-domain. The single-domain method further reduces the scale of the training graph and improves performance.
We conduct comprehensive evaluations and show that our model surpasses the state-of-the-art (SOTA) performance while saving space complexity by up to 10 times and training time by up to 3 times. Our code will be made public.

\begin{table}[t]
\centering
\begin{tabular}{lrrr} \toprule
\textbf{Domain} & \textbf{\# Files} & \textbf{\# Concepts} & \textbf{\# Pos. Relations}  \\\midrule
NLP &    1,717 & 322 &  1,551 \\
 CV      &  1,041   &  201        &  871   \\
 BIO     &  148    &   100       & 234    \\
\bottomrule
\end{tabular}
\vspace{3mm}
\caption{Statistics of the three domains from LectureBankCD \cite{li2021unsupervised}: Files (resource files: lecture slides); Pos. Relations (positive prerequisite relations).}
\label{tab:stats}

\end{table}

\section{Dataset and Task Definition}
The LectureBankCD \cite{li2021unsupervised} dataset consists of concepts, resources (lecture slides from top universities), and manually annotated prerequisite relations between concepts, in three domains: NLP, BIO and CV (computer vision). 
We show the statistics of the dataset in Table~\ref{tab:stats}. We follow the same experimental setting as \cite{li2021unsupervised}, treating NLP as the source domain and BIO and CV as target domains. 

We define cross-domain prerequisite chain learning as a binary classification problem. Given a source domain and a target domain, there are a number of concept pairs $(p,q)$ in each domain. The label for the concept pair $y$ is 1 if concept $p$ is a prerequisite of concept $q$ and 0 otherwise. We focus on the unsupervised transfer learning setting, in which the labels of the source domain $y_{src}$ are known, but those of the target domain $y_{tgt}$ are unknown. 


\section{Methodology}

\begin{figure}[t]
    \centering
    \includegraphics[width=0.7\textwidth]{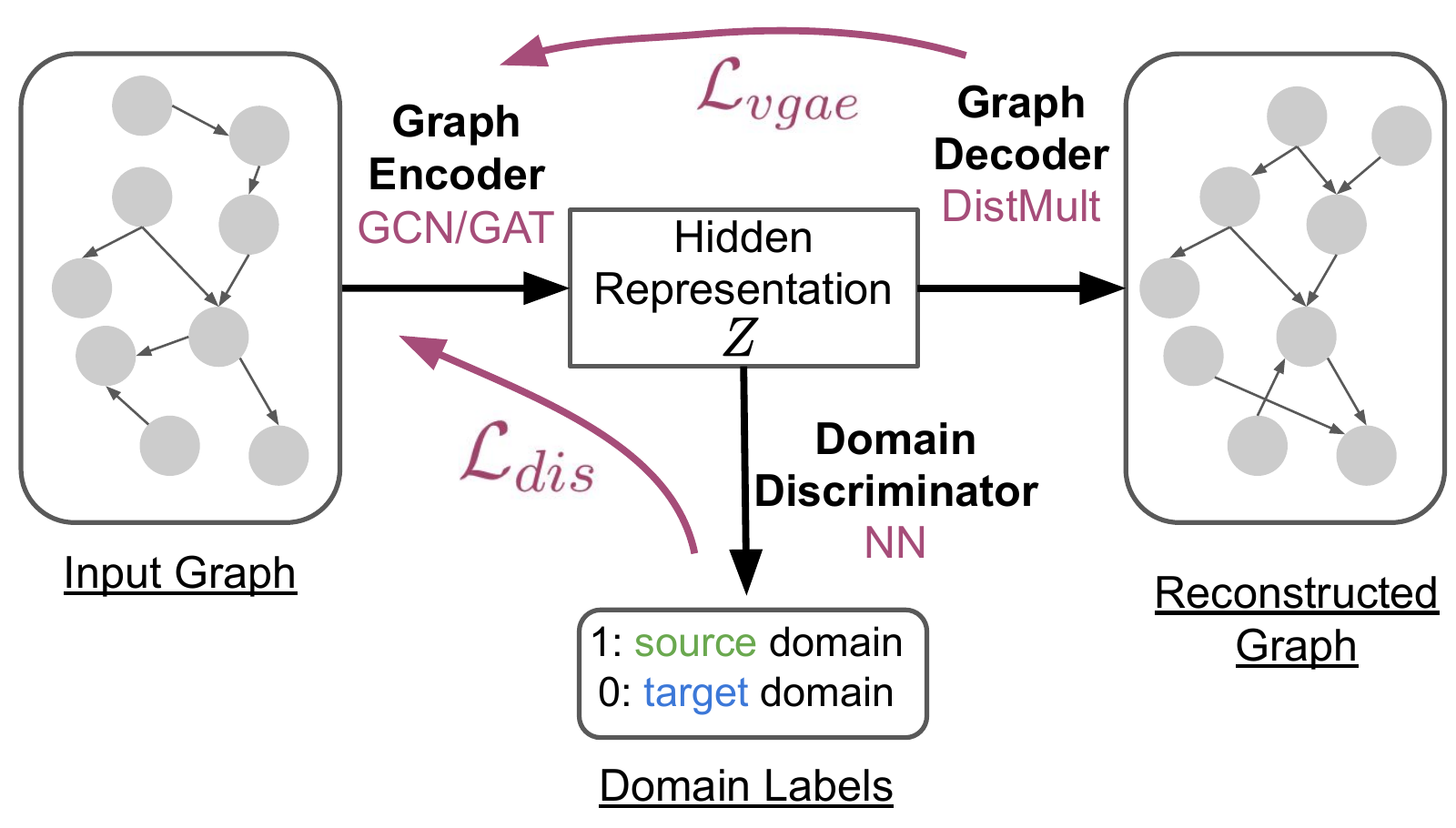}
    \caption{DAVGAE model.}
    \label{fig:adv_model}
\end{figure}

We propose the Domain Adversarial Variational Graph Autoencoders (DAVGAE) for unsupervised cross-domain prerequisite chain learning. The model architecture is shown in Figure~\ref{fig:adv_model}.

\textbf{Concept Graph Construction} 
We define a concept graph $G=(X,A)$ as the input to the model. $X$ is the set of node features and $A$ is the adjacency matrix which indicates whether prerequisite relations exist between concept pairs. If $p\to q$, we define $A_{p,q}=1$ and $A_{q,p}=1$. To obtain $X$, we follow the same approach from \cite{li2021unsupervised} to train Phrase2Vec (P2V) \cite{artetxe2018emnlp} node embeddings: we extract free text from LectureBankCD lecture slides then train a P2V model to encode concepts. 

We propose two ways to build the concept graph: \textbf{cross-domain} and \textbf{single-domain}. In the cross-domain one, all concept nodes are modeled in a single graph, and $X$ consists of concepts from the source and target domain.
We build the adjacency matrix $A$ using two information sources: relations between source domain concept nodes given in LectureBankCD during training, plus additional relations from cosine similarity or pairwise mutual information (PMI) of node embeddings between concept pairs. We calculate cosine similarity between all possible concept pairs, but only PMIs between source concepts and target concepts.

In the single-domain method, we train on two single-domain concept graphs to further reduce the space complexity during training: source graph $G_{src}$ and target graph $G_{tgt}$, in which $X$ of the two graphs only contains concept node features from the source and target domains respectively. In $G_{src}$, $A$ values consists of two parts: labeled relations between concept node pairs, and additional relations computed using cosine similarity. In contrast, the initial $A$ in $G_{tgt}$ only comes from cosine similarity.

\textbf{DAVGAE} The VGAE model \cite{kipf2016variational} contains a graph neural network (GCN) encoder \cite{kipf2016semi} and an inner product decoder. The loss of VGAE is defined as:

\begin{equation}
\begin{split}
\mathcal{L}_{vgae}=\mathbb{E}_{q(\mathbf{Z} \mid \mathbf{X}, \mathbf{A})}[\log p(\mathbf{A} \mid \mathbf{Z})]-\\
\mathrm{KL}[q(\mathbf{Z} \mid \mathbf{X}, \mathbf{A}) \| p(\mathbf{Z})],
\end{split}
\label{eq:loss}
\end{equation}
where the first term indicates the reconstruction loss, and the second term represents the KL Divergence between the hidden layer representation $Z$ and a normal distribution. It is possible to replace the GCN encoder with other architectures, such as graph attention networks (GAT) \cite{velickovic2018graph}. 

Domain-adversarial training is an established approach to learn representations for domain adaptation \cite{ganin2017domain}, but it has rarely been applied to graphs before to the best of our knowledge. To force the VGAE encoder to learn domain-invariant features of concept nodes, we add a \textit{domain discriminator} module to predict which domain each node in the hidden layer representation $Z$ belongs to. We use a two-layer neural network (NN) to predict domain labels: 1 if the node comes from the source domain and 0 otherwise. Thus, the discriminator loss $\mathcal{L}_{dis}$ is defined as a cross-entropy loss for domain prediction.
The total loss of the DAVGAE model is:
\begin{equation}
\mathcal{L} = \mathcal{L}_{vgae} + \mathcal{L}_{dis}
\end{equation}

We train DAVGAE at the graph level. In each epoch, we feed it with either the cross-domain graph or one of the single-domain concept graphs.

\textbf{Link Prediction} Since the prerequisite prediction should be asymmetric, it is not suitable to use an inner product decoder like the original VGAE did. Instead, we use DistMult \cite{yang2014embedding} to predict the link between a concept pair $(p,q)$ using a hidden layer representation $(Z_p,Z_q)$. Specifically, we reconstruct the adjacency matrix $\hat{A}$ by learning a trainable weight matrix $R$, such that $\hat{A}= Z^\intercal RZ$. Finally, we apply a Sigmoid function to determine positive/negative label on the value of $\hat{A}_{p,q}$.

\begin{table*}[t]
\centering
\footnotesize
\begin{tabular}{lccccccc}\toprule
  &  \multicolumn{3}{c}{NLP$\rightarrow$CV} & & \multicolumn{3}{c}{NLP$\rightarrow$BIO} \\ 
  \cline{2-4} \cline{6-8} 
 \textbf{Method}      & \textbf{F1}       & \textbf{Precision}    & \textbf{Recall}   &  & \textbf{F1}       & \textbf{Precision}    & \textbf{Recall}   \\ \midrule
\multicolumn{8}{l}{\textbf{\texttt{Unsupervised Baseline Models}}}  \\
   CLS + BERT   	& 0.4277    &	0.5743	& 0.3419   & &  0.3930         &  0.7481      &  0.2727      \\
   CLS + P2V   &   0.4881 	& 0.6106	& 0.4070 & & 0.2222  & 0.6000 & 0.1364 \\
GraphSAGE + P2V \tiny{\cite{graphsage17hamilton}}& 0.5342 & 0.5085 & 0.5515  &&  0.5283 &0.5177  & 0.5287 \\
   GraphSAGE + BERT \tiny{\cite{graphsage17hamilton}}& 0.5102 & 0.3611  & 0.5105  &&0.4736 & 0.4065 & 0.5180   \\
   VGAE + BERT \tiny{\cite{li2019whatsi}} & 0.5885 	 & 	0.5398 &	0.6488 &    &  0.6011        &     0.6185 &	0.5909\\
   VGAE + P2V \tiny{\cite{li2019whatsi}} & \underline{0.6202}   & 0.5368 & 0.7349 &&   \underline{0.6177} &     0.6521 &	0.6091       \\
\midrule	
\multicolumn{8}{l}{\textbf{\texttt{Baseline with Extra Resource Nodes}}} \\
CD-VGAE + BERT \tiny{\cite{li2021unsupervised}}   & 	0.6391  & 	0.5441 & 	0.7884   &&  0.6289 &    0.6425	& 0.6364     \\
    CD-VGAE + P2V   \tiny{\cite{li2021unsupervised}}  & \underline{0.6754}  & 0.5468 &	0.8837  &  & \underline{0.6512} & 0.6667 &	0.6364\\ 
\midrule 

\multicolumn{8}{l}{\textbf{\texttt{Cross-domain Concept Graph}} }  \\

GAT \cite{velickovic2018graph} &	0.6064 & 	0.5281&	0.7172&& 0.6257		&	0.5969	&	0.6609 \\
GAT + cos &	0.6276&	0.5276&	0.7793 && 0.6336		&	0.5644	&	0.7304 \\
GAT + cos + DAVGAE (\textbf{ours}) & 0.6251 &	0.5613 &	0.7218 && 0.6396	&	0.6557	&	0.6348 \\
\rowcolor{Gray}
GCN \cite{kipf2016semi} &	0.5951 &	0.5361&	0.6713 && 0.6319	&	0.6109	&	0.6609 \\
\rowcolor{Gray}
\rowcolor{Gray}
GCN + cos &	0.6318&	0.5379&	0.7655 && 0.6174	&	0.5991	&	0.6435 \\
\rowcolor{Gray}
*GCN + cos + DAVGAE (\textbf{ours}) &	\textbf{0.6321}&	0.5661  &	0.7195  && \textbf{0.6421}		&	0.5932	&	0.7130 \\

\midrule
  \multicolumn{8}{l}{\textbf{\texttt{Single-domain Concept Graph}} }  \\
 GAT \cite{velickovic2018graph} &		0.5573 &	0.4897&	0.7609 && 	0.5756&	0.5588&	0.6348 \\
GAT + cos &		0.6287&	0.5213&	0.8023 &&0.5587&	0.5248&	0.6261 \\
GAT + cos + DAVGAE (\textbf{ours})	&	0.6356	&0.5782&	0.7149 && 0.6545&	0.6024&	 0.7217 \\
 \rowcolor{Gray}
GCN \cite{kipf2016semi} 	&	0.5888&	0.5169	&0.6920 && 0.5304 & 0.5218&	0.6348 \\
\rowcolor{Gray}
GCN + cos&		0.6232	&0.5455	&0.7287 && 0.6117&	0.5599&	0.6783 \\

\rowcolor{Gray2}
*GCN + cos + DAVGAE (\textbf{ours}) & \textbf{0.6771} & 0.5734 & 0.8322 && \textbf{0.6738} &	0.6559	& 0.6957 \\


\bottomrule
\end{tabular}
\caption{Evaluation results on two target domains. Underlined scores are the best among the baseline models. }
\label{tab:res}
\end{table*}

\section{Evaluation}

We apply the same split on the data as two previous works \cite{li2019whatsi, li2021unsupervised}. Positive relations are divided into 85\% training, 5\% validation, and 10\% testing. Negative relations are sampled randomly to ensure balance between positive and negative relations. In Table \ref{tab:res}, we report average scores over five randomly seeded splits.

\textbf{Unsupervised Baseline Models}
We establish unsupervised baselines using both machine learning classifiers and graph embedding methods. For each method, we experiment with P2V \cite{artetxe2018emnlp} and BERT concept embeddings pretrained on our corpus. \texttt{CLS + BERT/P2V}: We adapt the Machine Learning baselines from an existing work \cite{li2021unsupervised}, concatenating paired concept embeddings and training a classifier.
\texttt{GraphSAGE+BERT/P2V}: We adapt GraphSAGE \cite{graphsage17hamilton} to generate node embeddings which are passed into DistMult. Model inputs include the BERT/ P2V embeddings of the source and target domain concepts, as well as an adjacency matrix constructed from annotations of source domain prerequisite relations and cosine similarities of target domain concept embeddings.
\texttt{VGAE+BERT/P2V}: We use a VGAE model  \cite{li2019whatsi} to predict concept pair relations. All baseline models are trained on the NLP domain and applied directly on the target domains, so we call them unsupervised baselines. 

\begin{table}[t]
\centering
\begin{tabular}{ccrr}\toprule
\textbf{Experiment} & \textbf{Model} & \textbf{\# Graph node} & \textbf{Computational time} \\ \midrule[0.5pt]
\multirow{2}{*}{\textbf{NLP$\to$CV}}&CD-VGAE & 3,281 & 127.5s\\
&Ours  & \textbf{322}& \textbf{47.1s} \\
\midrule[0.3pt]
\multirow{2}{*}{\textbf{NLP$\to$BIO}}&CD-VGAE & 2,287 & 71.6s\\
& Ours & \textbf{322} & \textbf{30.2s}\\
 \bottomrule
\hline
\end{tabular}
\vspace{3mm}
\caption{Comparison of graph scale and computation time. Computation time includes 200 epochs of training and one inference run. Ours: \texttt{GCN+cos+DAVGAE}.}
\label{tab:compare}
\end{table}

\textbf{Baseline with Extra Resource Nodes} The recent CD-VGAE model from \cite{li2021unsupervised} constructs a cross-domain concept-resource graph to predict target domain prerequisite relations via optimized VGAEs \cite{kipf2016variational}.

\textbf{Cross-domain, Single-domain Concept Graph} In both groups, we experiment with GAT and GCN (in light gray shaded color) graph encoders, as well as cosine similarity for additional edge values when building the input graph. GAT works better in some settings, but the best performing model is \texttt{GCN+cos+DAVGAE} in terms of F1 score, across both CV and BIO domains. The results suggest that DAVGAE does not require a complex graph encoder like GAT. A simple GCN is enough to train domain-invariant features. The same trend is observed when DAVGAE is trained on the single-domain concept graph setting. No matter whether we train on a cross-domain or single-domain concept graph, DAVGAE consistently improves upon comparable unsupervised baselines. Furthermore, a DAVGAE trained on a single-domain concept graph not only achieves better performance compared to one trained on a cross-domain concept graph, but it also outperforms CD-VGAE. Overall, our best performing model is \texttt{GCN+cos+DAVGAE} (marked with *). 

CD-VGAE is a competitive baseline, but it shows limited scalability as it trains on a larger graph and requires longer time for training. We list detailed numbers in Table \ref{tab:compare}. For example, in the NLP$\to$CV experiment, our best model is trained on a graph with 322 nodes while CD-VGAE constructs a large graph of 3,281 nodes. In the best case, DAVGAE requires only 10\% of the graph size and one-third of the training time as CD-VGAE does.

\section{Analysis}

We provide quantitative analysis and case studies on selected domains to shed light on the edges predicted by our model. 




\subsection{Quantitative Analysis} 
We compare our best model with the ground truth and another baseline model (\texttt{CLS+P2V}). We first recover the concept graph of the CV domain and review the degree of each concept node. Our model predicts 1,151 positive edges while the base model predicts 527. There are 871 in the ground truth. In general, our model has higher recall than the selected baseline. Higher recall is beneficial because we'd rather students learn extra concepts than miss important concepts.


\definecolor{color1}{RGB}{0,102,204}
\definecolor{color2}{RGB}{204,102,0}

\begin{table}[t]
\centering
\begin{tabularx}{0.93\textwidth} {ccX} \toprule
 \textbf{Domain} & \textbf{Graph} & \multicolumn{1}{c}{\textbf{Path}} \\ \midrule
 CV & Ground Truth & \textcolor{color1}{object recognition}, robotics, artificial intelligence,..., image processing, feature extraction, \textcolor{color2}{autonomous driving} \\
 & DAVGAE & \textcolor{color1}{object recognition}, video classification, \textcolor{color2}{autonomous driving} \\
\midrule[0.5pt]
 BIO &  Ground  Truth & \textcolor{color1}{DNA}, \textcolor{color2}{motif discovery}  \\
 & DAVGAE &  \textcolor{color1}{DNA}, dynamic programming, RNA secondary structure, energy minimization, decision trees, sampling, \textcolor{color2}{motif discovery} \\
 \bottomrule
\hline
\end{tabularx}
\vspace{3mm}
\caption{Case studies of concept paths.}
\label{tab:case-circles}

\end{table}

\subsection{Case Studies} 
In the concept graph recovered by DAVGAE, we observe that there are a few concept pairs which are connected by more than one path. The same is true for the ground truth graph. When there are cycles in the graph, finding all possible prerequisite paths becomes especially challenging. With this in mind, we conduct case studies on randomly selected paths.

In the CV domain, with random selection, there are usually 5-10 concepts in each path within the ground truth graph. The concept graph recovered by our model tends to have more and longer paths because more positive edges are predicted. In Table ~\ref{tab:case-circles}, we compare paths from the ground truth concept graph and our recovered graph. Both start with \textit{object recognition} (colored blue) and end with \textit{autonomous driving} (colored orange), with concepts linked from top to bottom. There exists a long path in the ground truth; however, but our model predicts a shorter one, indicating another possible learning path.
In BIO, we show paths from \textit{DNA} $\to$ \textit{motif discovery}. There are 8 paths found in the ground truth with an average path length of 4; in comparison, the 8 paths found in our model prediction has an average length of 10.63. We show the shortest path between the selected concepts in the ground truth and DVGAE concept graphs. Our model this time predicts more concepts along the path than the ground truth.

We include two random concept paths from CV and BIO in Table \ref{tab:casescv}. Both are predictions from our best performing model. In the left column, our model predicts many relations accurately in the CV domain, such as \textit{video classification}$\to$\textit{autonomous driving}, \textit{video and image augmentation}$\to$\textit{image generation} and \textit{image generation}$\to$\textit{image to image translation}. However, a few concepts may not be predicted in the correct path, e.g. \textit{gibbs sampling}. Similarly, in the right column, we observe correct prerequisite relations in the BIO domain, such as \textit{transcription}$\to$\textit{transcription factor} and \textit{ChIP-seq}$\to$\textit{gene finding}. 



\begin{table}[t]
\centering
\small
\begin{tabular}{c|c}\toprule
\textbf{CV} & \textbf{BIO} \\ \midrule
optical flow & BLAST \\
trajectory prediction &hardy-weinberg equilibrium\\
eye tracking & ChIP-seq\\
camera localization & gene finding\\
gibbs sampling & multivariate linear model\\
shading analysis & graph theory\\
background modeling and update & transcription\\
motion detection and tracking & transcription factor\\
action or gesture recognition & position weight matrix\\
video classification & yeast 2-hybrid\\
autonomous driving & energy minimization\\
remote sensing & markov clustering\\
crowd counting\\
graph rendering\\
image processing\\
video and image augmentation\\
image generation\\
image to image translation\\
 \bottomrule
\hline
\end{tabular}
\vspace{3mm}
\caption{A random concept path from CV (left column) and BIO  (right column).  }
\label{tab:casescv}
\end{table}

\section{Conclusion}

In this paper, we propose the DAVGAE model to solve cross-domain prerequisite chain learning efficiently. DAVGAE outperforms unsupervised baselines trained on concept graphs by a large margin. It also outperforms an unsupervised SOTA model trained on a concept-resource graph, while significantly reducing computation space and time.



\bibliography{custom}

\begin{thebibliography}{10}

\bibitem{alsaad2018mining}
F.~ALSaad, A.~Boughoula, C.~Geigle, H.~Sundaram, and C.~Zhai, ``Mining mooc
  lecture transcripts to construct concept dependency graphs.,'' {\em
  International Educational Data Mining Society}, 2018.

\bibitem{li2019whatsi}
I.~Li, A.~Fabbri, R.~Tung, and D.~Radev, ``What should {I} learn first:
  Introducing lecturebank for {NLP} education and prerequisite chain
  learning,'' in {\em The Thirty-Third {AAAI} Conference on Artificial
  Intelligence, {AAAI} 2019, The Thirty-First Innovative Applications of
  Artificial Intelligence Conference, {IAAI} 2019, The Ninth {AAAI} Symposium
  on Educational Advances in Artificial Intelligence, {EAAI} 2019, Honolulu,
  Hawaii, USA, January 27 - February 1, 2019}, pp.~6674--6681, {AAAI} Press,
  2019.

\bibitem{yu2020mooccube}
J.~Yu, G.~Luo, T.~Xiao, Q.~Zhong, Y.~Wang, W.~Feng, J.~Luo, C.~Wang, L.~Hou,
  J.~Li, Z.~Liu, and J.~Tang, ``{MOOCC}ube: A large-scale data repository for
  {NLP} applications in {MOOC}s,'' in {\em Proceedings of the 58th Annual
  Meeting of the Association for Computational Linguistics}, (Online),
  pp.~3135--3142, Association for Computational Linguistics, 2020.

\bibitem{yang2015concept}
Y.~Yang, H.~Liu, J.~G. Carbonell, and W.~Ma, ``Concept graph learning from
  educational data,'' in {\em Proceedings of the Eighth {ACM} International
  Conference on Web Search and Data Mining, {WSDM} 2015, Shanghai, China,
  February 2-6, 2015} (X.~Cheng, H.~Li, E.~Gabrilovich, and J.~Tang, eds.),
  pp.~159--168, {ACM}, 2015.

\bibitem{pan-etal-2017-prerequisite}
L.~Pan, C.~Li, J.~Li, and J.~Tang, ``Prerequisite relation learning for
  concepts in {MOOC}s,'' in {\em Proceedings of the 55th Annual Meeting of the
  Association for Computational Linguistics (Volume 1: Long Papers)},
  (Vancouver, Canada), pp.~1447--1456, Association for Computational
  Linguistics, 2017.

\bibitem{liang2018investigating}
C.~Liang, J.~Ye, S.~Wang, B.~Pursel, and C.~L. Giles, ``Investigating active
  learning for concept prerequisite learning,'' in {\em Proceedings of the
  Thirty-Second {AAAI} Conference on Artificial Intelligence, (AAAI-18), the
  30th innovative Applications of Artificial Intelligence (IAAI-18), and the
  8th {AAAI} Symposium on Educational Advances in Artificial Intelligence
  (EAAI-18), New Orleans, Louisiana, USA, February 2-7, 2018} (S.~A. McIlraith
  and K.~Q. Weinberger, eds.), pp.~7913--7919, {AAAI} Press, 2018.

\bibitem{li2021unsupervised}
I.~Li, V.~Yan, T.~Li, R.~Qu, and D.~Radev, ``Unsupervised cross-domain
  prerequisite chain learning using variational graph autoencoders,'' in {\em
  Proceedings of the 59th Annual Meeting of the Association for Computational
  Linguistics (ACL)}, 2021.

\bibitem{bojchevski2020scaling}
A.~Bojchevsi, J.~Klicpera, B.~Perozzi, A.~Kapoor, M.~Blais, B.~Rózemberczki,
  M.~Lukasik, and S.~Günnemann, ``Scaling graph neural networks with
  approximate pagerank,'' in {\em Proceedings of the 26th ACM SIGKDD Conference
  on Knowledge Discovery and Data Mining (KDD ’20)}, (Virtual Event, CA,
  USA), ACM, 2020.

\bibitem{frasca2020SIGN}
F.~Frasca, E.~Rossi, D.~Eynard, B.~Chamberlain, M.~Bronstein, and F.~Monti,
  ``Sign: Scalable inception graph neural networks,'' {\em arxiv}, 2020.

\bibitem{Goodfellow2014generative}
I.~J. Goodfellow, J.~Pouget{-}Abadie, M.~Mirza, B.~Xu, D.~Warde{-}Farley,
  S.~Ozair, A.~C. Courville, and Y.~Bengio, ``Generative adversarial
  networks,'' {\em CoRR}, vol.~abs/1406.2661, 2014.

\bibitem{chen2018adversarial}
X.~Chen, Y.~Sun, B.~Athiwaratkun, C.~Cardie, and K.~Weinberger, ``Adversarial
  deep averaging networks for cross-lingual sentiment classification,'' {\em
  Transactions of the Association for Computational Linguistics}, vol.~6,
  pp.~557--570, 2018.

\bibitem{chen2018multinomial}
X.~Chen and C.~Cardie, ``Multinomial adversarial networks for multi-domain text
  classification,'' in {\em Proceedings of the 2018 Conference of the North
  {A}merican Chapter of the Association for Computational Linguistics: Human
  Language Technologies, Volume 1 (Long Papers)}, (New Orleans, Louisiana),
  pp.~1226--1240, Association for Computational Linguistics, 2018.

\bibitem{li2021detecting}
I.~Li, ``Detecting bias in transfer learning approaches for text
  classification,'' {\em CoRR}, vol.~abs/2102.02114, 2021.

\bibitem{pan2018adversarially}
S.~Pan, R.~Hu, G.~Long, J.~Jiang, L.~Yao, and C.~Zhang, ``Adversarially
  regularized graph autoencoder for graph embedding,'' in {\em Proceedings of
  the Twenty-Seventh International Joint Conference on Artificial Intelligence,
  {IJCAI} 2018, July 13-19, 2018, Stockholm, Sweden} (J.~Lang, ed.),
  pp.~2609--2615, ijcai.org, 2018.

\bibitem{artetxe2018emnlp}
M.~Artetxe, G.~Labaka, and E.~Agirre, ``Unsupervised statistical machine
  translation,'' in {\em Proceedings of the 2018 Conference on Empirical
  Methods in Natural Language Processing}, (Brussels, Belgium), pp.~3632--3642,
  Association for Computational Linguistics, 2018.

\bibitem{kipf2016variational}
T.~N. Kipf and M.~Welling, ``Variational graph auto-encoders,'' {\em arXiv
  preprint arXiv:1611.07308}, 2016.

\bibitem{kipf2016semi}
T.~N. Kipf and M.~Welling, ``Semi-supervised classification with graph
  convolutional networks,'' in {\em 5th International Conference on Learning
  Representations, {ICLR} 2017, Toulon, France, April 24-26, 2017, Conference
  Track Proceedings}, OpenReview.net, 2017.

\bibitem{velickovic2018graph}
P.~Velickovic, G.~Cucurull, A.~Casanova, A.~Romero, P.~Li{\`{o}}, and
  Y.~Bengio, ``Graph attention networks,'' in {\em 6th International Conference
  on Learning Representations, {ICLR} 2018, Vancouver, BC, Canada, April 30 -
  May 3, 2018, Conference Track Proceedings}, OpenReview.net, 2018.

\bibitem{ganin2017domain}
Y.~Ganin, E.~Ustinova, H.~Ajakan, P.~Germain, H.~Larochelle, F.~Laviolette,
  M.~Marchand, and V.~S. Lempitsky, ``Domain-adversarial training of neural
  networks,'' in {\em Domain Adaptation in Computer Vision Applications}
  (G.~Csurka, ed.), Advances in Computer Vision and Pattern Recognition,
  pp.~189--209, Springer, 2017.

\bibitem{yang2014embedding}
B.~Yang, W.~Yih, X.~He, J.~Gao, and L.~Deng, ``Embedding entities and relations
  for learning and inference in knowledge bases,'' in {\em 3rd International
  Conference on Learning Representations, {ICLR} 2015, San Diego, CA, USA, May
  7-9, 2015, Conference Track Proceedings} (Y.~Bengio and Y.~LeCun, eds.),
  2015.

\bibitem{graphsage17hamilton}
W.~L. Hamilton, Z.~Ying, and J.~Leskovec, ``Inductive representation learning
  on large graphs,'' in {\em Advances in Neural Information Processing Systems
  30: Annual Conference on Neural Information Processing Systems 2017, December
  4-9, 2017, Long Beach, CA, {USA}} (I.~Guyon, U.~von Luxburg, S.~Bengio, H.~M.
  Wallach, R.~Fergus, S.~V.~N. Vishwanathan, and R.~Garnett, eds.),
  pp.~1024--1034, 2017.

\end{thebibliography}
\bibliographystyle{ieeetr}





\end{document}